\documentclass[10pt,twocolumn,letterpaper]{article}

\usepackage[pagenumbers]{cvpr} 

\usepackage{graphicx}
\usepackage{amsmath}
\usepackage{amssymb}
\usepackage{booktabs}
\usepackage{subfiles}
\usepackage{multirow}
\usepackage{stmaryrd}
\usepackage{soul}
\usepackage[dvipsnames]{xcolor}
\usepackage{tabularx}
\usepackage{bbding}
\usepackage{algorithm}
\usepackage{algpseudocode}
\usepackage{rotating}
\usepackage{longtable}
\usepackage{pgfplots}
\usepackage{pgfplotstable}
\usepackage{tikz}
\usepackage[colorinlistoftodos,prependcaption,textsize=tiny]{todonotes}
\graphicspath{ {./images/} }

\newcommand{\newref}[5]{
  \expandafter\newif\csname if#2first\endcsname
  \csname #2firsttrue\endcsname
  \expandafter\newcommand\csname #2\endcsname{%
    \csname if#2first\endcsname
      \textbf{#3}~\cite{#5}~\label{def:#2}%
      \csname #2firstfalse\endcsname
    \else
      \texorpdfstring{\hyperlink{cite.#5}}{#3}{#3}%
    \fi\xspace
  }%
  \ifx&#4&\else
    \expandafter\newcommand\csname #2short\endcsname{%
      \texorpdfstring{\hyperlink{cite.#5}}{#4}{#4}\xspace%
    }%
  \fi
}

\newref{model}{patchcore}{PatchCore}{PtC}{patchcore}
\newref{model}{reversedistillation}{RD4AD}{RD}{reverse_distillation}
\newref{model}{csflow}{CSFlow}{CSF}{csflow}
\newref{model}{msflow}{MSFlow}{MSF}{msflow}
\newref{model}{draem}{DRAEM}{DR}{draem}
\newref{model}{glass}{GLASS}{GL}{glass2024}
\newref{model}{simplenet}{SimpleNet}{SN}{simplenet}
\newref{model}{mmr}{MMR}{MMR}{mmr}
\newref{model}{inreach}{InReaCh}{InR}{mcintoshInReaCh}
\newref{model}{devnet}{DevNet}{DN}{pang2021explainable}
\newref{model}{dra}{DRA}{}{ding2022catching}
\newref{model}{efficientad}{EfficientAD}{}{efficientad}

\newref{dataset}{mvtec}{MVTecAD}{}{mvtec}
\newref{dataset}{visa}{VisA}{VA}{zou2022spot}
\newref{dataset}{btad}{BTAD}{}{btad}
\newref{dataset}{mpdd}{MPDD}{MP}{mpdd2020}
\newref{dataset}{riad}{Real-IAD}{}{riad}
\newref{dataset}{sensumds}{SensumSODF}{}{sensum_ds}
\newref{dataset}{aebad}{AeBAD}{}{mmr}
\newref{dataset}{vad}{VAD}{}{vad}
\newref{dataset}{viaduct}{VIADUCT}{}{viaduct}
\newref{dataset}{loco}{MVTecLOCO}{}{mvtecloco}

\newref{metric}{auroc}{AUROC}{}{auroc}
\newref{metric}{aupro}{AUPRO}{}{mvtec}

\definecolor{cvprblue}{rgb}{0.21,0.49,0.74}
\usepackage[pagebackref,breaklinks,colorlinks,allcolors=cvprblue]{hyperref}

\def\paperID{}
\def\confName{}
\def\confYear{2025}

\title{
    Beyond Academic Benchmarks: Critical Analysis and Best Practices for Visual Industrial Anomaly Detection
}

\author{
Aimira Baitieva$^1$, Yacine Bouaouni$^1$, Alexandre Briot$^1$, Dick Ameln$^2$, \\  Souhaiel Khalfaoui$^1$, Samet Akcay$^2$\\
\\
{$^1$Valeo, $^2$Intel}
\\
}

\begin{document}
\maketitle

\begin{abstract}
Anomaly detection (AD) is essential for automating visual inspection in manufacturing. This field of computer vision is rapidly evolving, with increasing attention towards real-world applications. 
Meanwhile, popular datasets are typically produced in controlled lab environments with artificially created defects, unable to capture the diversity of real production conditions. 
New methods often fail in production settings, showing significant performance degradation or requiring impractical computational resources. This disconnect between academic results and industrial viability threatens to misdirect visual anomaly detection research. This paper makes three key contributions: (1) we demonstrate the importance of real-world datasets and establish benchmarks using actual production data, (2) we provide a fair comparison of existing SOTA methods across diverse tasks by utilizing metrics that are valuable for practical applications, and (3) we present a comprehensive analysis of recent advancements in this field by discussing important challenges and new perspectives for bridging the academia-industry gap. The code is publicly available at 
\href{https://github.com/abc-125/viad-benchmark}{\textit{https://github.com/abc-125/viad-benchmark}}
\end{abstract}    
\section{Introduction}
\label{sec:intro}

Visual inspection is a quality control process in manufacturing that ensures products meet required standards and do not contain defects. This process is essential for cost reduction through early fault detection and preventing defective components from reaching customers. It is traditionally performed by human operators who suffer from fatigue and attention limitations \cite{human_errors}. Classical computer vision methods often lack the robustness and flexibility required for complex real-world scenarios \cite{liu2023deep_iad_survey, cv_for_vis_insp}.

\begin{figure}[htp]
    \centering
    \includegraphics[width=0.97\columnwidth]{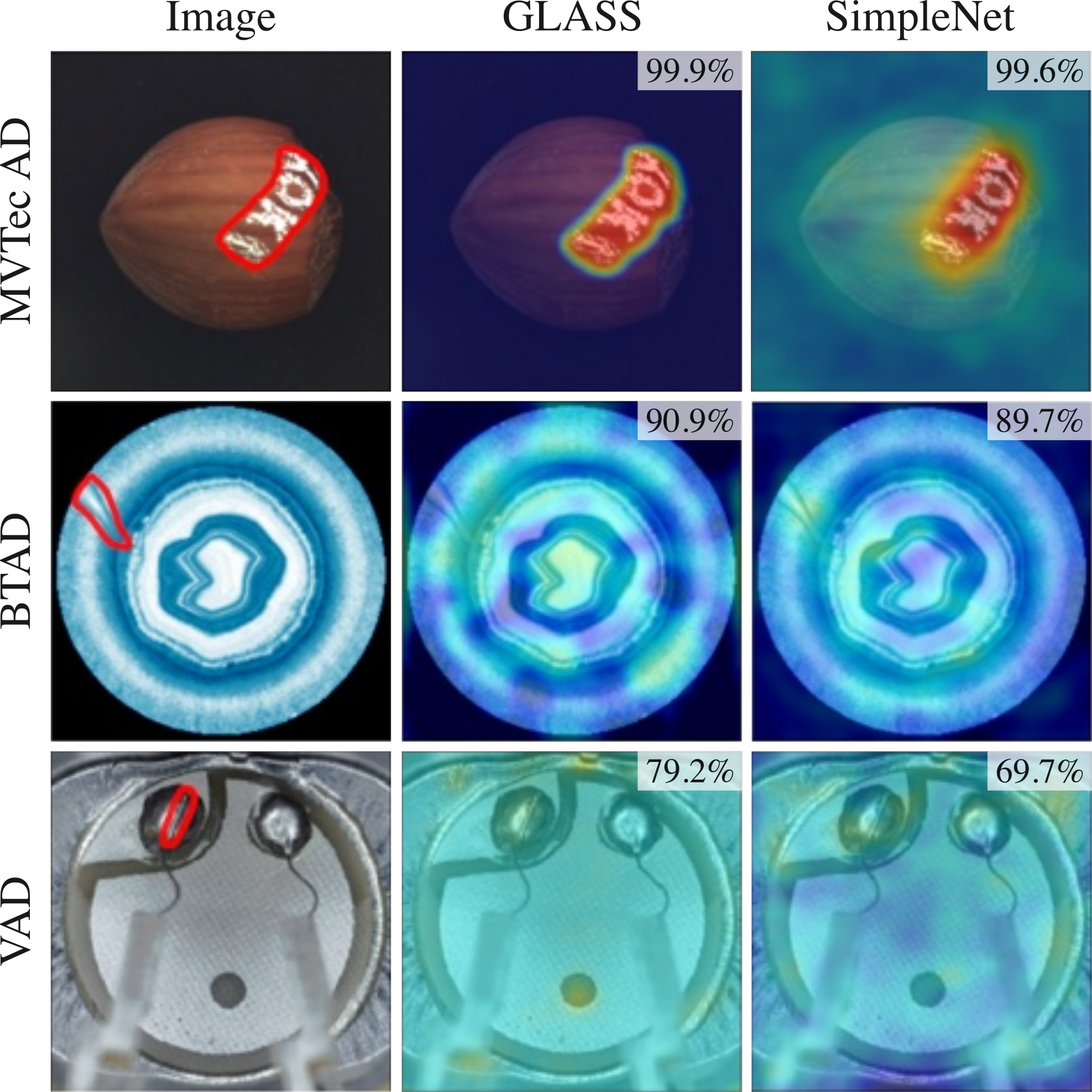}
    \caption{Anomaly maps generated by two recent SOTA models, GLASS \cite{glass2024} and SimpleNet \cite{simplenet}, overlayed over original images, demonstrate the reality gap in industrial anomaly detection. While the upper row illustrates the exceptional detections on MVTecAD dataset \cite{mvtec}, the next two rows display the examples that the models fail on real-world datasets, BTAD \cite{btad} and VAD \cite{vad}. Image-level AUROC for the whole dataset is shown in the top right corner of the respective anomaly map. More results can be found in Tab.~\ref{tab:general_results}.
    \label{fig:teaser}}
\end{figure}

Deep learning-based industrial AD has emerged as a promising solution. However, a significant gap exists between academic benchmarks and industrial deployment requirements, which results in large performance discrepancies in real production environments. Our analysis reveals several fundamental issues in this academic-industrial divide.
\begin{table}[ht]
\centering
\caption{Comparison of our paper and recent visual industrial AD benchmarks and surveys in terms of data and experiments.}
\label{tab:comparison}
\begin{tabularx}{\columnwidth}{c|X|cccc} 
 & Attribute & \cite{liu2023deep_iad_survey} & \cite{surveyad2024} & \cite{imiadbenchmark} & \textbf{Ours} \\
\hline
\multirow{2}{*}{\rotatebox{90}{ Data }} &
Number of classes & 15 & 44 & 52 & 65 \\
& Real-world classes & 0 & 0 & 10 & 25 \\
\hline
\multirow{4}{*}{\rotatebox{90}{ Experiments }} &
Input size & $\times$ & $\times$ & $\times$ & \Checkmark \\
& Noisy labels & $\times$ & $\times$ & \Checkmark & \Checkmark \\
& Data shift & $\times$ & $\times$ & $\times$ & \Checkmark \\
& Supervised AD & $\times$ & $\times$ & \Checkmark & \Checkmark \\
& Validation set & $\times$ & $\times$ & $\times$ & \Checkmark \\
\end{tabularx}
\end{table}

\paragraph{Datasets.}
A fundamental challenge in industrial anomaly detection is the lack of comprehensive datasets that contain real anomalous samples. This limitation stems from the difficulty of acquiring defective images in production environments \cite{viaduct, vad}, and causes unique challenges, such as the absence of validation splits and the inability to measure model performance on real-world data \cite{liu2023deep_iad_survey, surveyad2024}. While recent research primarily focuses on the \mvtec dataset, where models compete to achieve near-perfect image-level \auroc scores exceeding 99.9\% \cite{viaduct, glass2024}, such evaluations may not reflect real industrial challenges. To address this limitation, our work includes comprehensive experiments on four unresolved real-world datasets \cite{vad, viaduct, mmr, sensum_ds} to significantly expand the scope of industrial anomaly detection benchmarking, as shown in Tab.~\ref{tab:comparison}.

\vspace{-0.25cm}
\paragraph{Testing Setup.}
Our review of existing publications reveals critical limitations in current evaluation practices. The first issue is the widespread use of test set-based early stopping and selective reporting of best epoch results \cite{simplenet, msflow, yu2021fastflow}, which can lead to over-fitting and unrealistic performance estimates \cite{efficientad}. Second, is the use of center crop augmentation \cite{patchcore, cfa}, which can introduce biases in performance evaluation by reducing the number of false positives. This approach can also lead to missing defects outside the crop area in real-world applications. Our benchmark addresses these issues by implementing standardized evaluation protocols, such as using a fixed epoch count and disabling center crop for all methods to ensure more reliable comparisons. 

\vspace{-0.25cm}
\paragraph{Metrics.}
Most of the recent papers use the same image-level metric \auroc, which is not accountable for the relative importance of possible errors. In real production environments, misclassifying defective parts causes significantly higher costs than false positive classifications. To address this practical consideration, we include a metric used in the industry that focuses on classifying good parts correctly with a fixed percentage of defective parts missed. More details on this can be found in Subsec.~\ref{ssec:metrics}.  We also provide full results for four image-level and three pixel-level metrics; the link to the results can be found in the Supplementary. 

\vspace{-0.25cm}
\paragraph{Experiments.}
Based on extensive industry experience, we introduce five experiments to bridge the gap between academic research and industrial requirements: (1) investigating input size effects on detection performance; (2) evaluating model robustness under distribution shifts; (3) analyzing performance with noisy training data; (4) comparing supervised approaches on real-world datasets; and (5) investigating validation strategies to prevent test set over-fitting. We present detailed discussions in Sec.~\ref{sec:discussion}.

\section{Datasets}
\label{sec:datasets}

The collection of real-world defective product images is fundamentally challenging due to the scarcity of anomalies and commercial sensitivity concerns. For this reason, researchers have proposed a number of dataset creation strategies, from simple synthetic texture generation \cite{dagm2007} to collecting real objects and damaging them manually in the lab environment \cite{mvtec, zou2022spot}. While these controlled approaches helped immensely to develop new algorithms and push the state of industrial AD forward, they often fail to authentically replicate all the complex and diverse problems that can emerge on real production lines \cite{survey_industry_ad}. One example is that most recent datasets contain only 4-5 types of defects, while real-world ones might have more than 20 types \cite{vad}.

\begin{table}[ht]
\centering
\caption{Datasets included in this benchmark. Cls. = \textit{Number of classes}, Img. = \textit{Number of images}, Type = \textit{Type of defects}, lab = \textit{Defects are produced in the laboratory settings}, rw = \textit{Defects are produced in the real-world settings}, Log. def. = \textit{Logical defects}.
}
\label{tab:datasets}
\begin{tabularx}{\columnwidth}{l|XXXcc} 
Dataset & Year & Cls. & Img. & Type & Log. def. \\ 
\hline
BTAD \cite{btad}            & 2021 & 3  & 2830  & rw  & no \\ 
MPDD \cite{mpdd}            & 2021 & 6  & 1346  & rw  & no \\ 
SSODF \cite{sensum_ds}      & 2021 & 2  & 2180  & rw  & no \\ 
LOCO \cite{mvtecloco}       & 2022 & 5  & 3644  & lab & yes \\ 
VisA \cite{zou2022spot}     & 2022 & 12 & 10821 & lab & no \\ 
AeBAD-S \cite{mmr}          & 2023 & 1  & 2159  & rw  & no \\
VAD \cite{vad}              & 2024 & 1  & 5000  & rw  & yes \\
RIADs \cite{riad}           & 2024 & 8  & 8022  & lab  & no \\ 
VIADUCTs \cite{viaduct}     & 2024 & 12 & 2723  & rw  & yes \\
\end{tabularx}
\end{table}

\begin{table*}[ht]
\centering
\caption{Classification of AD models with respect to the data they require for training. Bold denotes models included in this benchmark.
}
\label{tab:models}
\begin{tabularx}{\textwidth}{|c|l|l|X|}
\hline
Unsupervised & \multicolumn{3}{l|}{\textbf{InReaCh} \cite{mcintoshInReaCh}, IGD \cite{igd2022}} \\ \hline
\multirow{3}{*}{One-class} & \multicolumn{3}{l|}{\textbf{PatchCore} \cite{patchcore}, \textbf{RD} \cite{reverse_distillation}, \textbf{CSFlow} \cite{csflow}, \textbf{MSFlow} \cite{msflow}, \textbf{SimpleNet} \cite{simplenet}, \textbf{MMR} \cite{mmr}, Padim \cite{padim}} \\ \cline{2-4}
                       & \multirow{2}{*}{Textures dataset} & \multicolumn{2}{l|}{\textbf{DRAEM} \cite{draem}, DeSTSeg \cite{zhang2023destseg}} \\ \cline{3-4}
                       & & Background masks & \textbf{GLASS} \cite{glass2024} \\ \hline
\multirow{2}{*}{Supervised} & \multicolumn{3}{l|}{\textbf{DevNet} \cite{pang2021explainable}, \textbf{DRA} \cite{ding2022catching}} \\ \cline{2-4}
                       & {Background masks, pixel-level labels} & \multicolumn{2}{l|}{BTAD \cite{btad}, PRN \cite{prnad}} \\ \cline{3-4}
                       \hline
\end{tabularx}
\end{table*}

This paper divides datasets into two groups, with defects produced in laboratory settings and real-world defects. It is important to note that sometimes, these groups cannot describe all the variety of data available and that real-world datasets are not necessarily more valuable. \cite{viaduct} state that they have both real-world inspection problems integrated and used as a blueprint. Some real-world datasets might have just several defective objects shot from different sides \cite{mpdd}, which may provide less valuable than datasets with hundreds of defects created in a lab \cite{riad} environment. We focus on datasets containing objects since texture-only datasets such as \cite{dagm2007, kolektorssd2019} are mostly solved and generally less challenging for modern deep learning methods \cite{assumptions_deep_ad}.

It is important to distinct between \emph{logical defects} (e.g., missing components, incorrect assembly, wrong shape of components) and \emph{structural defects} (e.g., cracks, pits, surface damage). Most of the emerging papers are trying to solve the problem of \emph{structural defects}. In general, to detect such defects, it is sufficient to identify abnormal features in the image. Logical defects, on the contrary, require detecting rather an absence of normal features, which is impossible without understanding the general context. We include datasets \cite{mvtecloco, vad, viaduct}, which contain logical defects in our analysis to show the importance of this problem.

\begin{figure}[h]
\centering
\includegraphics[width=\columnwidth]{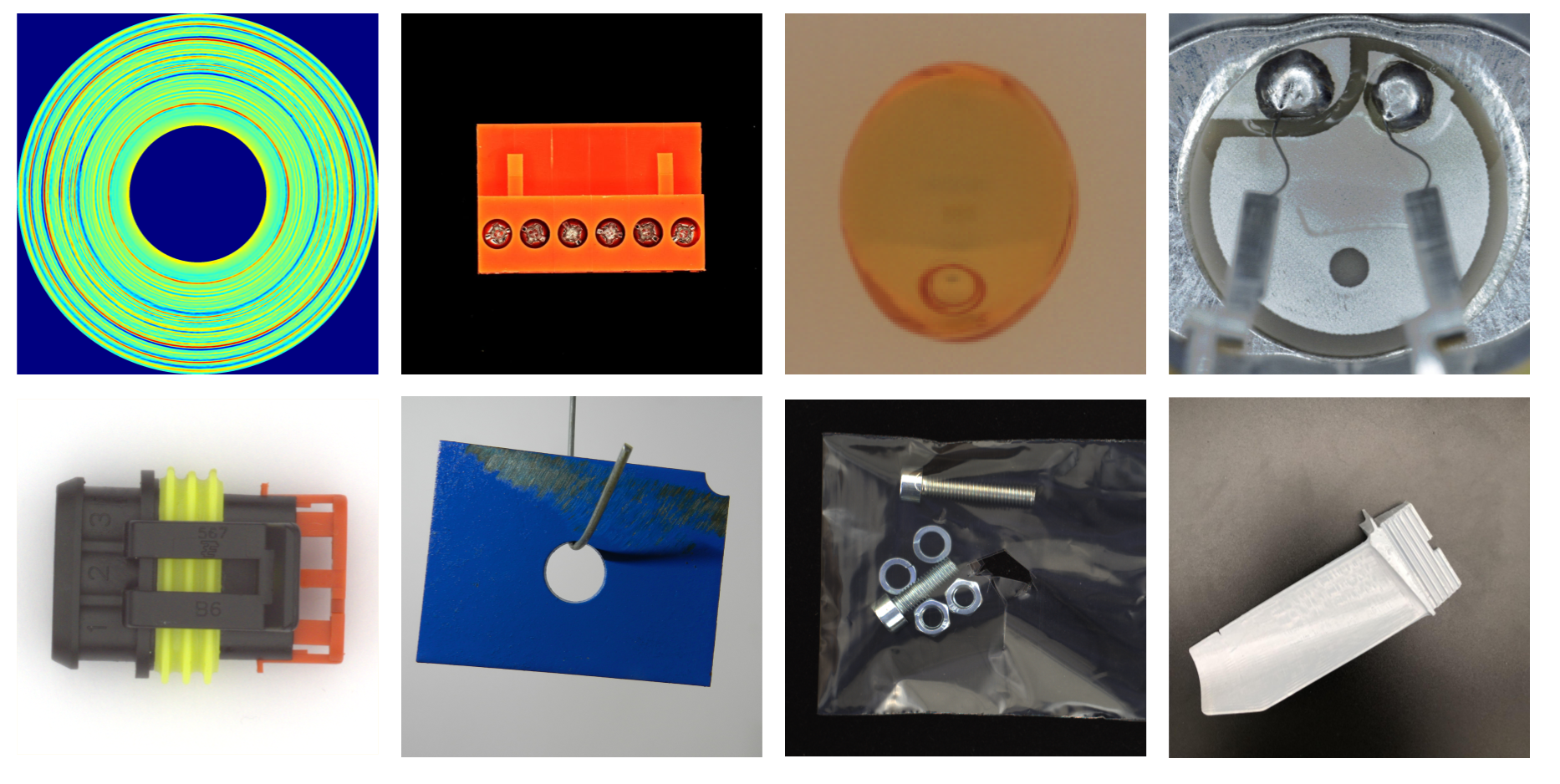}
    \caption{Examples of some of the object categories used in this benchmark.
    \label{fig:vis_datasets}}
\end{figure}

A description of the datasets can be found in Tab.~\ref{tab:datasets}, and a visualization of some categories is in Fig.~\ref{fig:vis_datasets}. Below, we describe details on several selected datasets. 
\riad is an industrial dataset with lab-made defects that provides 30 categories of objects. In our work, due to computational limitations and a more even representation of different datasets, we selected 8 hard and moderately hard categories, denoting this as Real-IAD-subset (RIADs). \btad is a dataset with images captured from the industrial acquisition systems. It contains 3 categories, one of which is especially challenging due to unusually large defects and high variability of good images. \sensumds is a real-world pharmaceutical inspection dataset with 2 categories of pills with diverse defects. \aebad presents photos of aero-engine blades with a domain shift between training and test sets. \vad is a one-category dataset with more than 20 types of real-world defects. \viaduct is the largest visual AD dataset available, with 49 categories comprising real-world defects or ones inspired by it. Similar to \cite{riad}, we select 12 categories and denote them as VIADUCT-subset (VIADUCTs). We do it to focus on including as many different datasets as possible rather than having just a few large datasets. Our aim is to capture diverse challenges, such as a low number of training images, multiple instances, backgrounds with false positive spots, different camera angles, different object complexity, and many others.

\section{Models}
\label{sec:models}

\begin{table*}[ht]
\centering
\caption{Evaluation of AD algorithms on 8 datasets using 4 metrics. The best result is marked in bold. Metrics are im.AUROC/im.PG2/ pix.AUPRO/pix.F1Max. VAD has no pixel-level results due to the absence of such labels.
\label{tab:general_results}}
\begin{tabularx}{\textwidth}{X|c|c|c|c|c|c|c|c} 
Method & BTAD    & SSODF    & RIADs      & VIADUCTs       & MPDD     & VisA & LOCO & VAD   \\ 
\hline
\patchcore  & \textbf{95.5}/67.3 & 93.2/45.0 & 91.4/39.0 & \textbf{86.5}/49.6 & 94.9/66.1 & 90.3/56.3 & 84.2/23.7 & \textbf{88.0}/16.5 \\
 & 76.9/55.1 & 92.7/52.8 & 92.0/33.3 & 90.2/30.7 & 89.2/42.2 & 86.3/42.7 & \textbf{76.7}/43.7 & —/— \\
\hline
\reversedistillation  & 94.3/67.7 & 91.9/36.3 & \textbf{93.2}/\textbf{46.3} & 83.1/43.0 & 93.3/61.5 & \textbf{95.0}/\textbf{67.4} & 77.0/13.4 & 84.7/20.1 \\
 & \textbf{79.5}/\textbf{58.5} & 92.4/55.7 & \textbf{95.0}/41.1 & \textbf{93.4/}32.3 & \textbf{93.2}/\textbf{43.4} & \textbf{91.5}/\textbf{44.9} & 75.9/38.8 & —/— \\
\hline
\mmr  & 93.7/62.9 & 91.1/39.0 & 92.4/42.7 & 80.9/31.4 & 93.0/54.8 & 93.2/57.3 & 84.2/23.7 & 87.6/27.6 \\
 & 77.9/57.4 & 92.9/55.8 & 96.3/40.6 & 87.4/28.2 & 94.8/50.3 & 89.4/43.4 & \textbf{76.7}/43.7 & —/— \\
\hline
\csflow  & 95.1/\textbf{71.5} & 91.2/37.3 & 86.3/22.7 & 82.4/43.7 & \textbf{96.8}/\textbf{82.8} & 90.9/55.8 & 81.6/\textbf{30.4} & 82.2/17.1 \\
 & 57.5/25.6 & 29.6/8.0 & 47.9/3.1 & 33.9/6.8 & 36.3/9.6 & 34.5/13.6 & 44.2/28.6 & —/— \\
\hline
\msflow  & 90.0/57.4 & 92.0/41.3 & 89.2/31.1 & 84.6/48.6 & 96.2/76.3 & 90.7/56.4 & 82.3/25.2 & 84.4/\textbf{26.6} \\
 & 62.6/32.3 & 77.5/28.0 & 85.6/8.9 & 73.4/15.2 & 60.0/16.3 & 80.6/24.6 & 71.0/38.7 & —/— \\
\hline
\simplenet  & 89.7/53.5 & 85.8/18.3 & 82.6/35.4 & 85.8/\textbf{53.5} & 86.6/53.3 & 90.8/46.5 & \textbf{84.3}/29.5 & 69.7/5.3 \\
 & 68.0/46.6 & 79.3/48.6 & 77.6/24.0 & 86.4/29.3 & 82.0/35.4 & 82.8/39.3 & 74.1/40.0 & —/— \\
\hline
\draem  & 89.6/55.9 & 86.1/17.7 & 85.6/26.9 & 79.8/29.9 & 94.8/73.2 & 89.0/54.1 & 75.3/19.1 & 57.7/1.8 \\
 & 52.6/25.2 & 83.4/42.5 & 86.8/\textbf{59.3} & 70.2/\textbf{35.5} & 74.7/32.3 & 69.5/35.8 & 65.5/\textbf{44.6} & —/— \\
\hline
\glass  & 90.9/45.5 & \textbf{95.5}/\textbf{54.7} & 88.8/30.7 & 76.9/33.8 & 90.1/60.8 & 94.2/59.5 & 80.7/27.4 & 79.2/10.3 \\
 & 61.1/49.6 & \textbf{95.2}/\textbf{57.6} & 67.1/38.8 & 73.5/24.2 & 89.3/37.9 & 86.0/39.2 & 69.5/41.9 & —/— \\
\hline
\end{tabularx}
\end{table*}

This work categorizes AD models based on their training data requirements, as summarized in Tab.~\ref{tab:models}. While existing taxonomies \cite{liu2023deep_iad_survey, surveyad2024, imiadbenchmark, adbench} focus on architectural aspects, our classification emphasizes practical deployment considerations. This is particularly critical with real-world industrial applications where pixel-level labels and/or background masks are often not feasible to obtain.

AD models are frequently characterized as unsupervised \cite{yu2021fastflow, pretrained_feature_extractors, liu2023deep_iad_survey, csflow}. However, this terminology is misleading because these models require only normal (one-class) images for training, which differs from the conventional definition of unsupervised models in other areas of computer vision \cite{dino, mae2021}. We propose the term \emph{one-class} to accurately describe these methods and distinguish them from truly \textit{unsupervised} approaches, which do not use any label information \cite{mcintoshInReaCh, igd2022}. \textit{Supervised} models, in contrast, utilize both good and bad classes for training.

While recent diffusion models show comparable results to methods we review in this paper, their substantial computational requirements make them impractical for production environments where rapid training and real-time inference are essential \cite{zhang2024realnet, yao2024glad, transfusion2024}. We focus on models that can train efficiently on consumer GPUs and meet strict production speed requirements. Unified multi-class models that can work with more than one category of object might solve this problem, but currently, their performance is worse than classic methods \cite{diff_multiclass2024, uniad}.
Zero-shot and few-shot VLMs, such as WinCLIP \cite{winclip}, AnoVL \cite{deng2024anovl}, InCTRL \cite{zhu2024toward}, and others \cite{aprilgan, gu2023anomalyagpt, li2024promptad, myriad}, use different training settings and fall out of the scope of this paper. However, they might be included in future work due to high practical interest.

\subsection{Unsupervised}
\label{ssec:unsupervised_models}
There are relatively few genuinely unsupervised AD vision models, which can be attributed to the inherent difficulty of detecting subtle deviations in a self-supervised manner instead of distinguishing between more distinct categories in other computer vision domains. Such models usually perform poorly compared to one-class models \cite{imiadbenchmark, mcintoshInReaCh, Qiu2022LatentOE}, so we select just one novel model for our benchmark for the noisy labels experiment. \inreach can operate with unlabeled data by rejecting anomalous images during training using only high-confidence normal patches across all samples. 

\subsection{One-class}
\label{ssec:oneclass_models}
One-class methods represent the predominant approach in AD. We evaluate several SOTA models based on different principles of describing a normal distribution by prioritizing approaches that balance detection performance with computational efficiency. \patchcore implements a memory bank approach that relies on a pre-trained feature extractor \cite{pretrained_feature_extractors} to distill features from the training set and compare them to the new data. It uses features from WideResNet \cite{wrn_clf}, levels 2 and 3, which improves its robustness to data drift but reduces its efficacy in detecting small defects. \reversedistillation is a teacher-student model where the student network restores the teacher's multiscale representations from high-level embeddings that enhance the diversity of anomalous representations. \csflow is a normalizing flow \cite{norm_flows} model that estimates the likelihood of the encoded features using multi-scale generative decoders and an EfficientNet \cite{efficientnet} as a pretrained encoder. \msflow is a similar normalizing flow model consisting of asymmetrical parallel flows and a fusion flow, which helps detect anomalies of different sizes. \draem is trained discriminatively to learn a decision boundary between normal and synthetic anomalous images. It requires the DTD dataset \cite{dtd2014} to synthesize defects, which might be less sufficient for complex defects. \glass is another discriminative model that synthesizes a broader range of anomalies at both the feature and image levels, improving the detection of defects close to normal regions. It also requires the usage of a texture dataset and background masks. \simplenet applies Gaussian noise at a feature level without external data sources; it uses a feature extractor similar to PatchCore and a discriminative classifier. \mmr employs masked multi-scale reconstruction task to enhance the model's ability to deduce causality among patches in normal samples, improving performance under data shifts. We exclude EfficientAD \cite{efficientad} from our study because it is patented, and its potential for commercial applications is highly limited.

\subsection{Supervised}
\label{ssec:supervised_models}
While supervised AD might theoretically offer advantages in boundary refinement \cite{deepsad}, practical limitations arise from the impossibility of capturing complete defect variability \cite{prnad}, the scarcity of anomalous training data, and the small size of anomalies in the image. Most existing approaches utilize only 10 randomly selected anomalous images, significantly diverging from real-world requirements. One exception is the \vad dataset, which contains 1000 anomalous images for training. 

Our comprehensive evaluation of supervised AD methods encountered notable reproducibility challenges. Despite thorough implementation attempts, we could not achieve the reported performance metrics for \cite{bgad}. Similarly, \cite{prnad} lacks publicly available official implementation, and existing community implementations demonstrate significant deviation from the reported results. These reproducibility considerations led us to focus our analysis on methods with verified implementations: \devnet, which leverages prior probability distributions to learn expressive representations of normality and unbounded deviations for anomalies, and \dra, which enhances detection through disentangled representations of seen anomalies, pseudo anomalies, and latent residual anomalies.

\section{Experiments}
\label{sec:experiments}

This paper includes experiments with single-class setting only \cite{patchcore, efficientad, csflow}, using one category of objects to train a model at a time. As our experiments show, this is already challenging for SOTA AD methods. For supervised AD, we follow the general setting \cite{zhu2024ahl, ding2022catching}, using a random selection from all defect types for training. Models we have tested performed insufficiently with this setting, so we do not include an even more challenging open-set setting.

InReaCh showed weak results in the Noisy labels experiment; for this reason, it is not included in the rest of the experiments. DRA and DevNet are included in the Supervised AD experiment only due to relatively low performance for the number of defective training images required.

In most of the tables, we use short names for models due to size constraints and easier readability than citation numbers. These names are constant across tables and contain clickable links to citations. InR means \textit{InReaCh}, PtC \textit{PatchCore}, RD \textit{RD4AD}, CSF \textit{CSFlow}, MSF \textit{MSFlow}, SN \textit{SimpleNet}, DR \textit{DRAEM}, GL \textit{GLASS}, DN \textit{DevNet}.

\subsection{Metrics}
\label{ssec:metrics}
To comprehensively evaluate anomaly detection performance in industrial settings, we employ seven distinct metrics: four image-level (\textit{denoted by im.}), and three pixel-level (\textit{denoted by pix.}), implemented in Anomalib library \cite{anomalib}. Image-level metrics assess whole-image classification accuracy, while pixel-level metrics evaluate the precision of defect localization. Our selection of metrics balances traditional academic evaluation criteria with practical industrial requirements.

The Area Under the Receiver Operating Characteristic curve, \textbf{AUROC} \cite{auroc}, measures the classification performance across multiple thresholds, which is valuable for image-level tasks. However, it is less suitable for tasks with an extreme class imbalance (such as pixel-level segmentation, where most pixels are normal) \cite{Saito2015ThePP, Rafiei2023OnPP}. The more informative pixel-level metric is the Area Under the Per-Region Overlap curve, \aupro, which measures recall for anomalous regions, disregarding thresholds that produce False Positive Rate (FPR) values exceeding 30\% when calculating the area under the PRO curve \cite{bertoldo2024aupimo}. 

To better align with industrial requirements, we use Presorted Good at 2\% (\textbf{PG2}), which shows the percentage of correctly classified good parts at 2\% of bad parts classified incorrectly. This metric emphasizes the practical application of AD models by showing their potential to reduce human operator workload while maintaining an acceptable level of misclassification for bad parts. It can be interpreted as True Negative Rate (TNR) at 2\% False Negative Rate (FNR), assuming bad parts are considered the positive class. Presorted Bad at 2\% (\textbf{PB2}) shows the percentage of bad parts classified correctly at 2\% of good parts misclassified. Both these metrics are image-level. This metric was originally used in \cite{ind_framework} to describe the performance of the industrial AD framework, which included several models.

The \textbf{F1-Max} score is the maximum F1 Score that can be achieved by varying the threshold of a binary classifier. F1 Score is computed as the harmonic average between precision and recall, representing the best balance between precision and recall for any single threshold value. This makes it a useful measure of a model’s performance in practical applications, where the final classification labels are more important than the raw anomaly scores. The F1-Max score has been used to rate both image-level and pixel-level performance of anomaly detection models \cite{winclip, zou2022spot}.

\subsection{Unified setup}
\label{ssec:setup}
To ensure fair comparison and reproducibility, we establish a standardized evaluation framework. All images are processed at $256\times256$ pixels resolution (except in the input size experiment, see Subsec.~\ref{ssec:input_size}). Following Batzner \etal \cite{efficientad}, we eliminate center crop augmentation, which may improve performance on commonly used datasets, but relies on the potentially dangerous assumption that defects never occur in peripheral regions. To prevent test set leakage, we avoid using test data for early stopping or model selection.
For architectural consistency, we employ WideResNet50 as the feature extractor across all models where possible (except \csflow, \dra, and \devnet, due to implementation constraints). This ensures a fair comparison across different models without utilizing the extra performance that larger feature extraction models might yield \cite{pretrained_feature_extractors, yu2021fastflow}.

\subsection{Model Evaluation}
\label{ssec:general_results}
We present a comprehensive evaluation across multiple datasets in Tab.~\ref{tab:general_results}. The \aebad dataset is analyzed separately in our data drift study due to its emphasis on real-world shifts in the distribution of test images. For \vad, we report only image-level metrics due to the absence of pixel-level annotations. 

\subsection{Impact of Input Resolution}
\label{ssec:input_size}

\begin{figure}[h]
\centering
\includegraphics[width=\columnwidth]{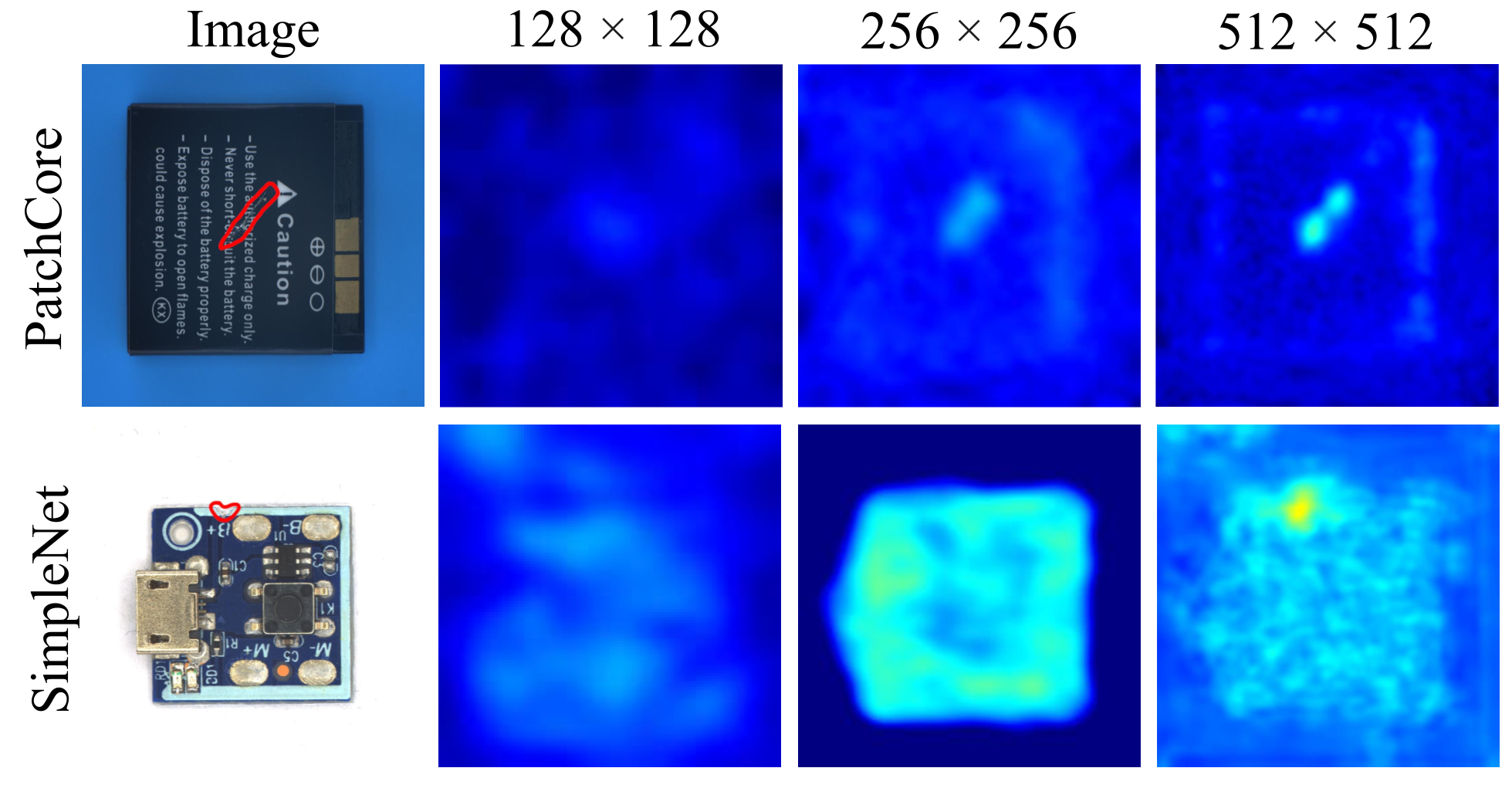}
    \caption{Anomaly maps generated by PatchCore and SimpleNet for Real-IAD, phone\_battery and pcb for different input sizes. The defect is circled in red on the original image.
    \label{fig:vis_input_size}}
\end{figure}

While increasing input resolution intuitively suggests improved performance \cite{patchcore, msflow}, the relationship between image size and detection accuracy requires careful examination. We analyze three resolution scales ($128\times128$, $256\times256$, and $512\times512$ pixels) across RIADs, BTAD, and VAD datasets (12 categories total). These datasets were selected for their high-resolution source images and diverse defect characteristics. Real-IAD's small defects provide particularly valuable insights into resolution-dependent detection capabilities, as demonstrated in Fig.~\ref{fig:vis_input_size}.

\begin{table}[htp]
\centering
\caption{Input size experiment. The best result is marked in bold. D denotes \textit{the mean value for RIADs, BTAD, VAD}. M means \textit{Methods}. Metrics are im.AUROC/pix.F1Max/im.PG2.
\label{tab:input_size}}
\begin{tabularx}{\columnwidth}{l|X|X|X} 
M & D128           & D256           & D512           \\ 
\hline
\patchcoreshort  & \textbf{88.3}/30.3/33.4 & \textbf{91.6}/44.2/40.9 & \textbf{94.1}/\textbf{57.3}/\textbf{60.0} \\
\reversedistillationshort   & 87.2/37.8/\textbf{35.0} & 90.7/\textbf{49.8}/\textbf{44.7} & 82.5/55.9/41.3 \\
\mmrshort & 81.7/33.2/19.8 & 91.2/49.0/44.4 & 88.9/52.9/37.2 \\
\csflowshort   & 81.6/12.1/21.8 & 87.9/14.3/37.1 & 89.6/17.1/44.0 \\
\msflowshort  & 81.2/9.2/29.0 & 87.9/20.6/38.4 & 86.0/33.4/40.2 \\
\simplenetshort  & 76.8/27.2/25.9 & 80.7/35.3/31.4 & 81.1/39.6/33.1 \\
\draemshort   & 80.5\textbf{/42.6}/26.8 & 77.6/42.2/28.2 & 78.3/34.9/15.9 \\
\glassshort   & 77.4/28.3/20.6 & 86.3/44.2/28.8 & 89.2/53.0/37.1
\end{tabularx}
\end{table}

\subsection{Robustness to Data Drift}
\label{ssec:data_drift}

\begin{table}[htp]
\centering
\caption{Data drift experiment. The best result is marked in bold. D denotes \textit{the mean value for RIADs, BTAD, VAD}. M means \textit{Methods}. Metrics are im.AUROC/pix.AUPRO.
\label{tab:data_drift}}
\begin{tabularx}{\columnwidth}{l|X|X|X|X} 
M    & D         & D+dr      & AeBAD      & AeBAD+dr           \\ 
\hline
\patchcoreshort             & \textbf{91.6}/84.5 & \textbf{74.4}/\textbf{29.9} & 71.9/83.2 & 70.3/82.1 \\
\reversedistillationshort   & 90.7/\textbf{87.3} & 48.6/18.5 & 80.7/89.2 & 77.0/86.9 \\
\mmrshort                   & 91.2/87.1 & 65.2/21.1 & \textbf{84.8}/\textbf{92.4} & \textbf{81.8}/\textbf{89.5} \\
\csflowshort                & 87.9/52.7 & 59.4/15.7 & 63.7/33.4 & 56.3/29.4 \\
\msflowshort                & 87.9/74.1 & 53.6/3.5 & 62.2/70.0 & 50.7/36.9 \\
\simplenetshort             & 80.7/72.8 & 56.5/15.6 & 60.4/76.7 & 52.1/66.9 \\
\draemshort                 & 77.6/69.7 & 52.4/13.0 & 59.6/70.4 & 54.8/44.0 \\
\glassshort                 & 86.3/64.1 & 55.9/11.8 & 72.7/86.5 & 55.6/81.6 \\
\end{tabularx}
\end{table}

Production environments frequently experience variations in imaging conditions that can significantly impact model performance. It is, therefore, important to develop models that are robust to such changes. We evaluate robustness to distribution shift through two complementary approaches:
\vspace{-0.25cm}
\paragraph{Synthetic Perturbations.} For the RIADs, BTAD, and VAD datasets, we simulate real-world variations through augmentations such as Gaussian noise or Gaussian blur, salt and pepper noise, perspective transformation, color jitter, and random shadowing. We apply these augmentations using three seeds and report the average results. Details of the exact methodology used to obtain the perturbations are provided in the Appendix~\ref{subsec:synth_data_drift}.
\vspace{-0.25cm}
\paragraph{Natural Perturbations.} We use the AeBAD dataset to evaluate the performance under real-world data drift, such as background, lighting, and camera angle, which were produced by changing actual shooting conditions \cite{mmr}. We report the average results for these types of drift in the column AeBAD+dr in Tab.~\ref{tab:data_drift}. Results without drift are displayed in column AeBAD.

\begin{figure}[h]
\centering
\includegraphics[width=\columnwidth]{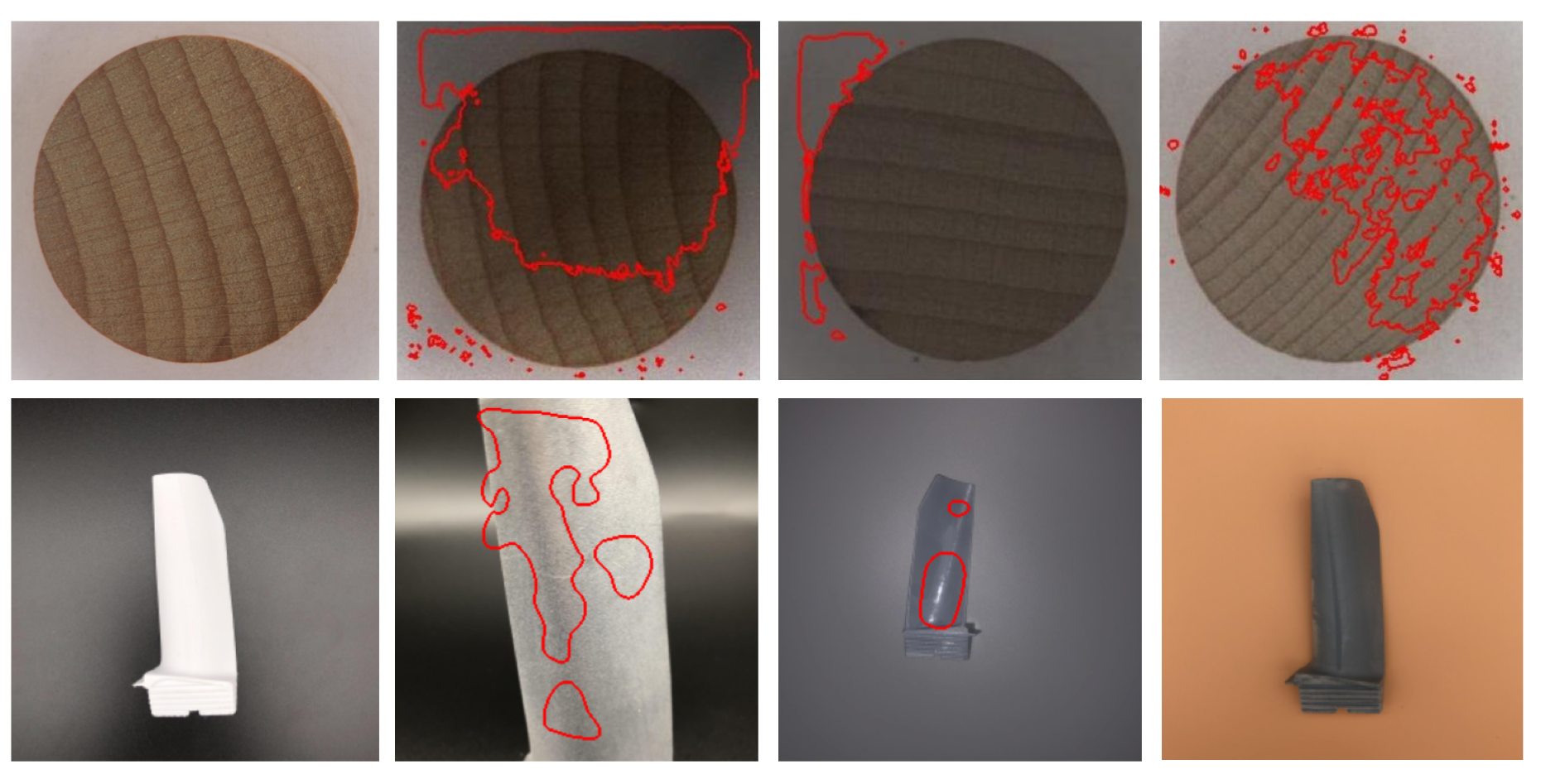}
    \caption{Pixel-level predictions generated by DRAEM for Real-IAD, woodstick, and SimpleNet for AeBAD, overlay over original images. All images contain no defects. Upper row, from left to right: no synthetic perturbations, random shadowing, color jitter, added noise. Lower row: no natural perturbations, change of camera angle, different lighting, different background.
    \label{fig:vis_data_drift}}
\end{figure}

\subsection{Robustness to Noisy labels}
\label{ssec:noisy_labels}
The problem of training one-class models is that they require labeled data, which can be difficult to obtain under real-world conditions. This would be due to (i) the cost of manual labor that limits the resources for labeling tasks and (ii) human operator's susceptibility to missing certain defect types up to 25\%, which, overall, causes contamination of the training data \cite{human_errors}. In this study, we evaluate model robustness by 4\%, 8\%, and 16\% of the training set (normal images) being replaced by the same number of anomalous images from the test set. We analyze 15 categories across VAD, SensumSODF, and VIADUCTs datasets, which extends results of \cite{imiadbenchmark} to more challenging real-world scenarios. Results are averaged across three random seeds for the contamination procedure and training.

\begin{table}[htp]
\centering
\caption{Noisy labels experiment. The best result is marked in bold. D denotes \textit{the mean value for VAD, SensumSODF, and VIADUCTs datasets}. M means \textit{Methods}. Metrics are im.AUROC/pix.AUPRO.
\label{tab:noisy_labels}}
\begin{tabularx}{\columnwidth}{l|X|X|X|X}
M    & D         & D4\%      & D8\%      & D16\%           \\
\hline
\inreachshort & 71.6/75.4 & 78.9/77.4 & 79.7/77.0 & 78.5/77.4 \\
\patchcoreshort  & \textbf{89.2}/91.4 & \textbf{87.3}/81.9 & \textbf{86.3}/81.1 & \textbf{84.7}/79.7 \\
\reversedistillationshort  & 87.5\textbf{/92.9} & 85.4/\textbf{91.5} & 85.0/\textbf{92.0} & 84.0/\textbf{92.0} \\
\mmrshort & 86.5/90.1 & 84.9/90.2 & 84.1/90.3 & 83.3/90.7 \\
\csflowshort  & 85.3/31.7 & 83.7/37.3 & 81.9/38.0 & 80.9/39.8 \\
\msflowshort  & 87.0/75.4 & 85.5/78.0 & 84.0/79.8 & 81.9/81.5 \\
\simplenetshort  & 80.4/82.9 & 76.2/79.3 & 72.2/73.5 & 68.8/70.5 \\
\draemshort   & 74.6/76.8 & 71.7/65.1 & 71.7/61.0 & 68.6/61.4 \\

\end{tabularx}
\end{table}

\subsection{Supervised AD}
\label{ssec:sup}
As discussed in Sec.~\ref{sec:models}, the scarcity of supervised anomaly detection datasets has limited progress in this direction. Most studies utilize 10 randomly selected anomalous images from the test set to the training set and evaluate results across multiple seeds \cite{zhu2024ahl, ding2022catching}. However, based on results in \cite{vad}, 10 images are insufficient for more complex datasets. For this reason, we use a higher number for our experiments: 20 for VIADUCTs and 50 for SensumSODF, based on the number of available anomalous images in the test set. For VAD, we utilize all 1000 bad training images from the training set. We report average results over three seeds.

\begin{table}[htp]
\centering
\caption{Supervised AD experiment. The best result is marked in bold. M means \textit{Methods}. Metrics are im.AUROC/im.F1Max/im.PG2.
\label{tab:sup}}
\begin{tabularx}{\columnwidth}{l|X|X|X} 
M & VIADUCTs           & SSODF           & VAD           \\ 
\hline
\dra & \textbf{85.1}/\textbf{82.0}/\textbf{45.2} & \textbf{87.8}/\textbf{80.9}/\textbf{26.4} & 88.6/80.9/\textbf{21.3} \\
\devnetshort & 77.0/69.6/36.9 & 86.6/80.1/20.8 & \textbf{89.0}/\textbf{83.7}/12.5 \\
\end{tabularx}
\end{table}

\subsection{Validation Strategies}
\label{ssec:val_set}

As discussed in Sec. ~\ref{sec:intro}, the widespread practice of using test sets for early stopping in recent anomaly detection models introduces potential over-fitting and unreliable performance estimates. To address this methodological issue, we conduct experiments on VAD dataset that contains both normal and anomalous samples in the training set. We use a 10\% validation split of the training data to avoid any test set leakage over three seeds.

\begin{table}[htp]
\centering
\caption{Validation set experiment on VAD dataset. M means \textit{Methods}.
Metrics are im.AUROC/im.F1Max.
\label{tab:val_set}}
\begin{tabularx}{\columnwidth}{l|X|c|X|c} 
M    & No early stop         & Epochs      & Early stop      & Epochs           \\ 
\hline
\csflowshort  & 81.8/74.6 & 240 & 84.5/76.8 & 65 \\
\reversedistillationshort  & 84.4/76.9 & 200 & 86.6/78.4 & 35 \\
\simplenetshort  & 68.4/68.3 & 160 & 82.5/74.9 & 12 \\
\end{tabularx}
\end{table}

\section{Discussion}
\label{sec:discussion}

\subsection{Problems and perspectives}
\label{ssec:dis1}

\paragraph{Real-world datasets.}
The introduction of the \mvtec dataset in 2019 marked a significant advancement in industrial AD. This dataset provided images with authentic defects and objects, distinguishing from earlier datasets that primarily featured textures and simple defects \cite{dagm2007, kolektorssd2019, kol2}. As of 2024, the field is showing signs of benchmark saturation for this dataset, with image-level AUROC scores exceeding 99.9\%. This raises questions about the potential overfitting of models to \mvtec and the generalization of performance comparisons to other benchmarks. Our findings point out that various state-of-the-art models \cite{simplenet, glass2024} significantly underperform older models \cite{patchcore, reverse_distillation} when evaluated on different datasets \cite{vad, viaduct, btad, riad}, which confirms the notion that the field is biased toward \mvtec. These findings underline the importance of evaluating models across diverse datasets, including real-world ones, to mitigate the risk of overfitting to a single dataset. It is also critically important to collect additional real-world datasets from actual production lines, such as \cite{btad, vad}. These datasets can present unique challenges that differ from those encountered in real-world multiview datasets such as \cite{mpdd, mmr}. 

\vspace{-0.25cm}
\paragraph{Generating defects.}
Although defect generation has advanced significantly over the last several years \cite{cutpaste, nsa_synt_anomalies}, generating realistic logical defects remains challenging. Such an effort will require a deep understanding of logical constraints, which can be difficult even for humans with their excessive world knowledge. As shown in Tab.~\ref{tab:general_results}, discriminative methods \cite{draem, simplenet, glass2024}, which rely heavily on the generation of defects, fail on more complex datasets such as \vad and \viaduct. The same problem appears in some supervised AD methods \cite{ding2022catching}. Novel methods, such as RealNet \cite{zhang2024realnet}, use diffusion to generate more realistic defects \cite{diff_generation}. But they still perform the best on rather simple datasets and structural defects and impose practical limitations of the high computational cost and lengthy training process \cite{zhang2024realnet}.

\paragraph{Results and further research.}
Our input resolution experiment shows that improvements depend heavily on the model architecture. In some cases, performance may even deteriorate at higher resolutions, which we attribute to the varying utilization of feature extractor layers and their receptive fields \cite{patchcore, pretrained_feature_extractors}. In other cases, higher resolution allowed to achieve significantly better results with 1.7-2.9 image-level AUROC improvement.

With EfficientAD's patenting, the field currently lacks freely available models capable of detecting logical defects at production-grade levels, which represents another gap between academic research and industrial needs \cite{ast2023, studentteacher, fanogan}.

The data drift experiment shows that current approaches are prone to changes in data distribution. This property stems from their basic design principle of detecting minimal deviations from training data, which can be problematic in dynamic production environments. While the transformer-based \cite{vit} models show promise in handling global context \cite{Pirnay2021InpaintingTF, comad}, they struggle with small defect detection compared to CNNs \cite{comad}. \patchcore's superior performance under distribution shift appears linked to the fact it uses layer2 and 3 of ResNet, avoiding layer1, which might be a source of a problem for \reversedistillation and \msflow. Another interesting outcome of this experiment is that most AD models could be used to detect data drift.

Our validation experiment highlights the need for robust protocols that reflect real-world deployment conditions. The challenge lies in developing appropriate metrics and optimization criteria based solely on normal samples, as anomalous samples are often limited or unavailable during initial deployment. Another possible solution might be to restrict defects used for validation to just one type at once, following the open-set supervised AD paradigm \cite{zhu2024ahl}.

The noisy labels experiment demonstrates that although one-class models are commonly called unsupervised, their performance fell significantly with a higher proportion of mislabeled images. Truly unsupervised approaches, such as \inreach, still do not outperform most one-class methods, even with a high proportion of noisy labels. Their average performance remains too unreliable for industrial deployment, reinforcing the importance of properly labeled training data in production environments.

Intuitively, using defective images for training should improve results significantly. The supervised AD experiment demonstrates otherwise. 
Such results can be explained by how some supervised AD models work: they generate artificial anomalies in addition to real ones, and such anomalies might belong to a different distribution than the test data. Overfitting architecture and parameters to \mvtec could be another reason.

The Appendix provides further discussion of results and different metrics, visualizations of image-level score distributions, and results per dataset.

\subsection{Best practices}
\label{ssec:dis2}

\paragraph{Datasets.}

Collecting real-world data, particularly anomalous images, is crucial to industrial AD. Although progress has been made in this direction, significant efforts are still required. Datasets with objects occupying most of the image are still limited, although such a setup more accurately reflects real-world challenges and conditions. Eliminating excessive background is often achievable even with traditional computer vision techniques. Another important aspect is the inclusion of validation splits, ideally encompassing both normal and anomalous samples.

\vspace{-0.25cm}
\paragraph{Models.}
The evaluation of anomaly detection models requires significant standardization to better reflect real-world requirements. Our results clearly show that success on \mvtec has become insufficient for predicting deployment readiness, as top-performer models often struggle with other datasets. This shows how including diverse, complex, and unresolved datasets is important. Recent papers might include \visa, which is a step in the correct direction, although more complex datasets such as \loco are still not widely used. In addition to using multiple datasets, overfitting the test set through early stopping or selective epoch reporting should be eliminated. This is critical as it creates dangerous illusions of progress that do not translate to real-world applications. Using fixed-epoch or fixed-iteration count, removing center crop augmentation, and maintaining consistent input processing across comparisons should ideally become the norm. These seemingly minor methodological choices can significantly impact performance on real-world industrial data. In addition, comprehensive metric reporting is essential since different types of errors carry different costs in production environments, and our evaluation practices should reflect this.

\section{Conclusion}
\label{sec:conclusion}

This paper provides an in-depth analysis of visual industrial AD to highlight critical gaps between academic benchmarks and industrial requirements. Our extensive experiments across nine datasets, eleven SOTA models, and seven metrics reveal that current evaluation practices may not reliably predict deployment success, as models with 99.9\% image-level AUROC on \mvtec show significant degradation on real-world data. We performed five experiments inspired by practical challenges, addressing various types of anomaly detection tasks and ensuring a robust model evaluation. Our findings demonstrate that standardized evaluation practices and real-world performance metrics are essential for bridging the academia-industry gap. Future work may involve conducting additional experiments on a broader range of models and datasets, evaluating computational efficiency trade-offs, and developing robust validation strategies based only on non-defective images.

\clearpage
{
    \small
    \bibliographystyle{ieeenat_fullname}
    \bibliography{main}
}

\maketitlesupplementary
\setcounter{page}{1}

\section{Appendices}
\label{sec:appendices}

The full results for all methods and categories can be found in this file: \url{https://eu.mydrive.ch/shares/88698/f808671f908f1e93d2a5597c4cd29e9c/download/452956104-1740916171/results.json}

\begin{itemize}
    \item \textbf{Sec.~\ref{subsec:models}: Models} - Implementation details for models.
    \item \textbf{Sec.~\ref{subsec:datasets}: Datasets} - Implementation details for datasets.
    \item \textbf{Sec.~\ref{subsec:synth_data_drift}: Synthetic data drift} - Detailed descriptions of synthetic perturbations used to create synthetic data drift.
    \item \textbf{Sec.~\ref{subsec:score_distr}: Score distributions} - Visualization of image-level score distributions.
    \item \textbf{Sec.~\ref{subsec:more_results}: More results} - PG2 and PB2 image-level results for data drift and noisy labels experiments. Results for experiments per dataset.
\end{itemize}

\subsection{Models}
\label{subsec:models}

This subsection describes the details of the implementation of the models we use. All models use the pretrained feature extractor WideResNet-50-2 by TorchVision \cite{paszke2017automatic} unless stated otherwise. Input size is set to $256 \times 256$ for all experiments except Input Resolution. We remove center crop augmentation and report results for the last epoch without an early stop (unless it is a Validation Strategies experiment). The rest of the settings we use follow the original paper's settings.

\textbf{PatchCore}: Implementation by Anomalib~\cite{anomalib}. The feature extractor uses layers 2 and 3. The coreset sampling ratio is 0.1, and the number of neighbors is 9. For larger datasets (such as VAD and one of the classes in BTAD), for resolution $512 \times 512$, we sample 25\% of images for training to fit into GPU memory, similar to \cite{bao2022miad}. 

\textbf{Reverse Distillation}: Implementation by Anomalib~\cite{anomalib}. Trained for 200 epochs.

\textbf{CSFlow}: Implementation by Anomalib~\cite{anomalib}. EfficientNet B5 by TorchVision \cite{paszke2017automatic} as a feature extractor. Trained for 240 epochs.

\textbf{MMR}: Official implementation\footnote{\url{https://github.com/zhangzilongc/MMR}} re-implemented into Anomalib. Trained for 200 epochs with 50 warmup epochs.

\textbf{MSFlow}: Official implementation\footnote{\url{https://github.com/cool-xuan/msflow}}. The original code calculates two anomaly maps, one using addition to calculate pixel-level AUROC and another using multiplication to calculate pixel-level AUPRO. We modified it to calculate only one anomaly map through the addition; the same anomaly map is used to calculate the anomaly score, according to the paper.

\textbf{DRAEM}: Official implementation\footnote{\url{https://github.com/VitjanZ/DRAEM}}. Trained for 700 epochs.

\textbf{SimpleNet}: Official implementation\footnote{\url{https://github.com/DonaldRR/SimpleNet}} re-implemented into Anomalib. Trained for 160 epochs.

\textbf{GLASS}: Official implementation\footnote{\url{https://github.com/cqylunlun/GLASS}}. For VAD and BTAD, we do not use background masks because there is no background in the images. For the rest of the datasets, masks are created using SAM \cite{kirillov2023segany}. Trained for 640 epochs.

\textbf{DRA}: Official implementation \cite{ding2022catching}. This model calculates only image-level anomaly scores. We use the backbone ResNet-18 because other backbones have not been implemented. Trained for 30 epochs. We removed random rotation augmentation because it makes results for some datasets worse \cite{vad}.

\textbf{DevNet}: Official implementation \cite{pang2019deep}. This model calculates pixel-level anomaly scores separately from image-level scores based on gradient back-propagation, we do not include this in our evaluation. Trained for 50 epochs.

\subsection{Datasets}
\label{subsec:datasets}

This subsection describes the details of the datasets we use. Datasets not mentioned below are used without any changes, similar to the original papers. For \textbf{MVTecLOCO}, we report mean results for logical and structural defects, similar to the original paper. For \textbf{AeBAD}, we report mean results for three types of data drift and results without data drift separately.

\textbf{SensumSODF}: The original paper uses three splits; we use one split to simplify evaluation.

\textbf{Real-IAD}: We use a version of the dataset with a shot from above only. Classes included into our subset: \textit{pcb, phone\_battery, plastic\_nut, plastic\_plug, porcelain\_doll, terminalblock, usb, woodstick}.

\textbf{VIADUCT}: Classes included into our subset: \textit{3\_pole\_socket\_housing, cylinder\_screw, damper\_large, dsub\_connector, mains\_tester, pcb, retractor, ring\_cable\_lug, tack, terminal\_block\_a, threaded\_fitting, valve\_handle\_blue}.

\subsection{Synthetic data drift}
\label{subsec:synth_data_drift}

\begin{table*}
    \centering
    \caption{Overview of the transforms used in the data drift experiment. Each test set image was augmented with 1 or 2 transforms picked randomly from different categories.}
    \label{tab:augmentations}
    \begin{tabularx}{\textwidth}{llll}
    Category               & Transform             & Parameter          & Value/Range{*} \\
    \hline
    Motion /camera quality & Gaussian Blur         & Kernel size        & 7            \\ 
                           &                       & Sigma              & [0.1, 1.5]   \\
                           & Gaussian Noise        & Mean               & 0.5          \\ 
                           &                       & Standard Deviation & 1.0          \\ 
                           &                       & Scale              & [0.01, 0.05] \\
    \hline
    Lighting conditions    & Color Jitter          & Brightness factor  & [0.5, 1.5]   \\
                           &                       & Contrast factor    & [0.5, 1.5]  \\
                           &                       & Saturation factor  & [0.5, 1.5]   \\
                           & Random Shadow         & Number of Shadow Layers   & 3            \\
                           &                       & Brightness factor  & 0            \\
    \hline
    Camera position        & Random Rotation       & Rotation angle     & [-5, 5]     \\
                           & Random Cropping       & Scale              & [0.8, 1.0]   \\
                           & Perspective Transform & Distortion scale   & 0.2          \\
    \hline
    \multicolumn{4}{l}{\footnotesize * The used parameter value was sampled uniformly from the specified range of values} \\
    \end{tabularx}
\end{table*}

\begin{figure}[h]
\centering
\includegraphics[width=\columnwidth]{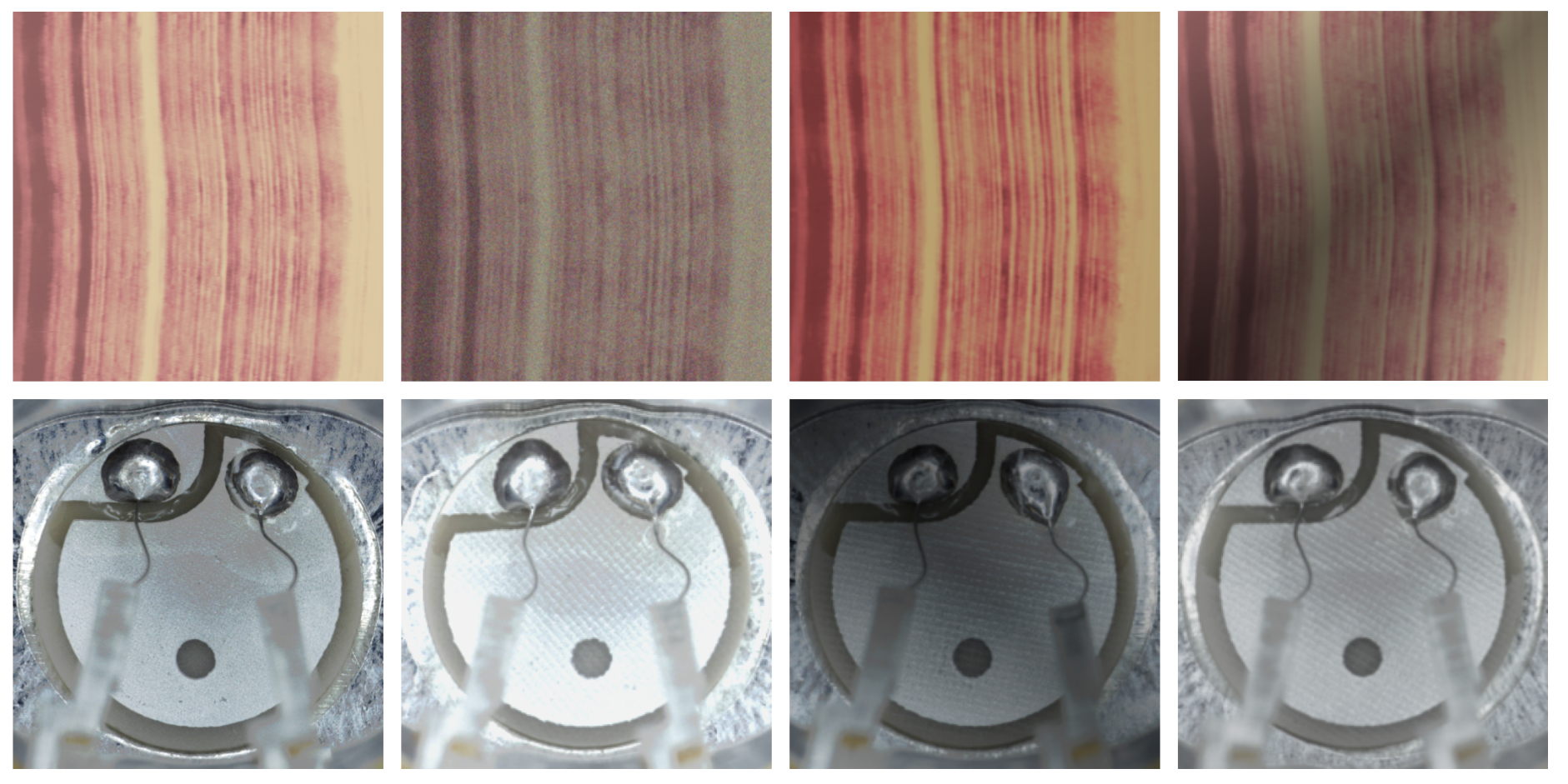}
    \caption{Synthetic perturbations. The image on the left in each row shows the original data, and the images on the right show different augmentations produced using our pipeline.
    \label{fig:data_drift_appendix}}
\end{figure}

To simulate data drift, an augmentation pipeline was created to synthetically add data drift to the images of the test set. The data drift augmentations consisted of 7 types of transforms across 3 categories, listed in Table \ref{tab:augmentations}. Gaussian blur and Gaussian noise to simulate camera motion and camera quality, color jitter and random shadow to simulate varying lighting conditions, and random rotation, random cropping and perspective transforms to simulate varying camera placement conditions. A random selection of 1 or 2 data drift categories was applied to each image from the test set. Within each of the selected categories, a single transform type was chosen randomly.

\subsection{Score distributions}
\label{subsec:score_distr}

\begin{figure*}
\centering
\includegraphics[width=\textwidth]{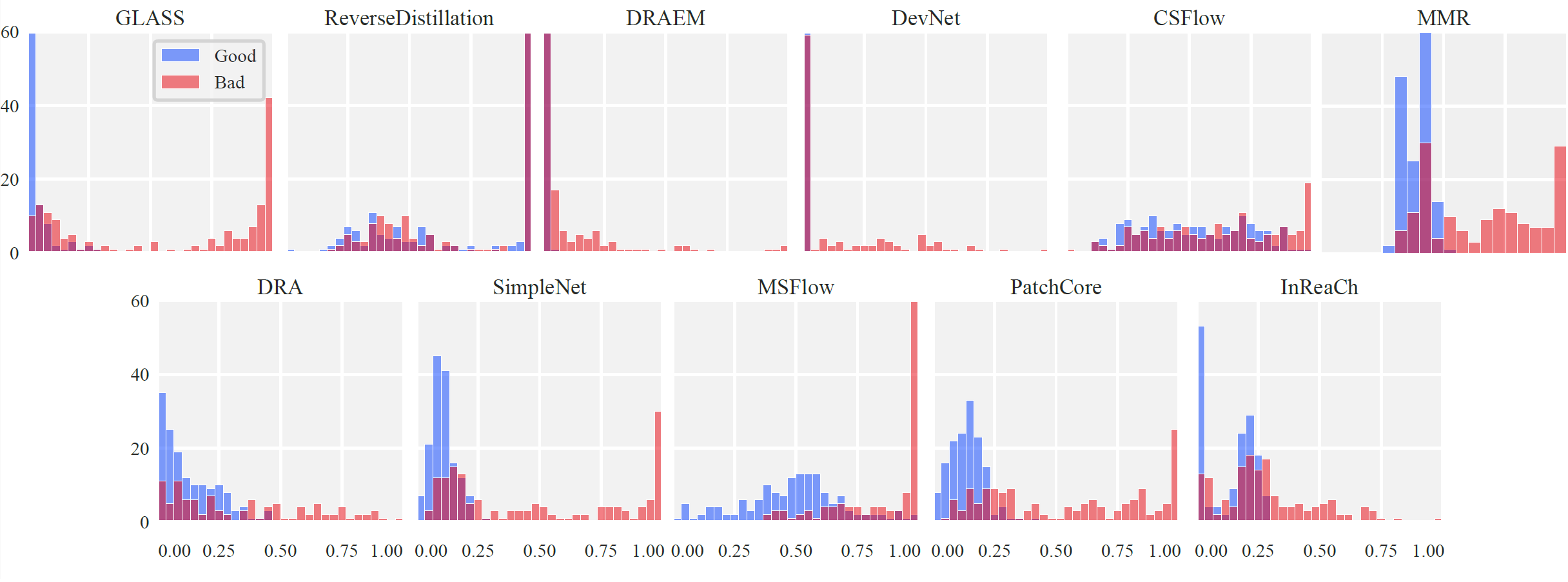}
    \caption{Image-level score distributions for different models for SensumSODF, capsule. Bad part scores are shown in red, and good in blue. The X axis shows scores, and the Y axis shows frequency.
    \label{fig:vis_histograms}}
\end{figure*}

\subsection{More results}
\label{subsec:more_results}

\begin{table}[htp]
    \centering
    \caption{Noisy labels experiment, additional results. The best result is marked in bold. D denotes \textit{the mean value for VAD, SensumSODF, and VIADUCTs datasets}. M means \textit{Methods}. Metrics are im.PG2/im.PB2.
\label{tab:noisy_labels_pgpb}}
\begin{tabularx}{\columnwidth}{l|X|X|X|X}
M    & D         & D4\%      & D8\%      & D16\%           \\
\hline
    \inreachshort  & 15.9/28.2 & 18.7/60.4 & 18.3/61.2 & 19.1/58.8 \\
    \patchcoreshort  & 37.0/56.9 & 30.0/87.1 & 33.1/81.4 & 26.8/79.5 \\
    \reversedistillationshort  & 33.1/48.9 & 30.0/71.7 & 30.9/65.2 & 30.6/66.8 \\
    \mmr  & 32.7/49.8 & 28.3/69.6 & 27.1/66.9 & 27.1/64.4 \\
    \csflowshort  & 32.7/47.4 & 31.1/63.4 & 26.2/57.8 & 26.0/55.5 \\
    \msflowshort  & 38.8/43.9 & 36.2/49.1 & 32.4/45.4 & 28.0/37.9 \\
    \simplenetshort  & 25.7/42.9 & 18.5/44.1 & 15.8/38.6 & 14.0/28.7 \\
    \draemshort  & 16.4/34.8 & 14.0/46.8 & 11.2/34.9 & 14.9/27.5 \\
\end{tabularx}
\end{table}

\begin{table}[htp]
\centering
\caption{Data drift experiment, additional results. The best result is marked in bold. D denotes \textit{the mean value for RIADs, BTAD, VAD}. M means \textit{Methods}. Metrics are im.PG2/im.PB2.
    \label{tab:data_drift_pgpb}}
\begin{tabularx}{\columnwidth}{l|X|X|X|X}
    M    & D         & D+dr      & AeBAD      & AeBAD+dr           \\
    \hline
    \patchcoreshort  & 40.9/66.5 & 14.2/25.3 & 11.7/15.3 & 15.2/13.2 \\
    \reversedistillationshort  & 44.7/63.8 & 2.0/0.8 & 38.7/22.2 & 31.7/16.3 \\
    \mmr  & 44.4/63.5 & 17.3/5.4 & 26.1/42.0 & 23.7/27.8 \\
    \csflowshort  & 37.1/54.3 & 3.6/4.1 & 9.1/5.9 & 6.9/3.3 \\
    \msflowshort  & 38.4/41.2 & 5.2/0.0 & 10.9/0.0 & 3.8/0.0 \\
    \simplenetshort  & 31.4/50.3 & 3.4/0.9 & 2.2/0.0 & 3.9/0.0 \\
    \draemshort  & 28.2/43.1 & 3.4/2.9 & 7.8/0.0 & 4.4/0.0 \\
    \glassshort  & 28.8/49.4 & 5.8/1.5 & 15.7/0.0 & 4.9/0.0 \\
\end{tabularx}
\end{table}

Due to restricted space in the paper, we include PG2 and PB2 image-level results for Noisy Labels and Data Drift experiments in Tab.~\ref{tab:noisy_labels_pgpb} and Tab.~\ref{tab:data_drift_pgpb} respectively. It can be seen that the best-performing models in terms of PG2 are different compared to image-level AUROC. In the Data Drift experiment, MSFlow performs better than PatchCore, which is also demonstrated in Fig.~\ref{fig:vis_histograms}, which shows a clearer separation of good images (measured by PG2) by MSFlow versus PatchCore. With the PG2 metric, PatchCore clearly outperforms other models; it is visualized in Fig.~\ref{fig:vis_histograms}. 

\begin{table}[htp]
\centering
\caption{Noisy labels experiment per dataset. Metrics are im.AUROC/pix.AUPRO.
    \label{tab:noisy_labels_per_dataset}}
    \begin{tabularx}{\columnwidth}{l|l|X|X|X}
    Method & Dataset & D4\%      & D8\%      & D16\%           \\
\hline
\inreachshort & SSODF  & 77.0/69.4 & 80.7/70.1 & 77.7/70.2 \\
 & VIADUCTs  & 76.6/85.4 & 75.3/83.9 & 75.2/84.7 \\
 & VAD  & 83.1/0.0 & 83.0/0.0 & 82.8/0.0 \\
\hline
\patchcoreshort & SSODF  & 92.6/91.0 & 92.6/90.9 & 90.7/89.7 \\
 & VIADUCTs  & 81.8/72.8 & 80.9/71.3 & 78.8/69.7 \\
 & VAD  & 87.4/0.0 & 85.6/0.0 & 84.8/0.0 \\
\hline
\reversedistillationshort & SSODF  & 91.0/93.4 & 91.1/93.7 & 89.4/93.8 \\
  & VIADUCTs  & 81.5/89.7 & 81.0/90.4 & 79.7/90.2 \\
  & VAD  & 83.8/0.0 & 82.8/0.0 & 82.9/0.0 \\
\hline
\mmr & SSODF  & 90.8/93.1 & 90.1/93.4 & 89.6/94.5 \\
  & VIADUCTs  & 79.5/87.3 & 78.8/87.2 & 78.1/86.8 \\
  & VAD  & 84.3/0.0 & 83.3/0.0 & 82.1/0.0 \\
\hline
\csflowshort & SSODF  & 89.5/40.0 & 88.1/42.4 & 88.0/44.4 \\
  & VIADUCTs  & 81.1/34.6 & 78.7/33.6 & 76.5/35.2 \\
  & VAD  & 80.3/0.0 & 78.8/0.0 & 78.0/0.0 \\
\hline
\msflowshort & SSODF  & 90.1/82.0 & 88.9/84.8 & 86.9/87.9 \\
  & VIADUCTs  & 82.9/74.1 & 80.9/74.8 & 78.4/75.0 \\
  & VAD  & 83.5/0.0 & 82.2/0.0 & 80.4/0.0 \\
\hline
\simplenetshort & SSODF  & 80.0/71.8 & 74.1/60.9 & 70.0/56.9 \\
  & VIADUCTs  & 84.1/86.7 & 81.3/86.0 & 79.9/84.1 \\
  & VAD  & 64.3/0.0 & 61.2/0.0 & 56.6/0.0 \\
\hline
\draemshort & SSODF  & 83.9/66.7 & 82.1/60.4 & 78.6/62.5 \\
 & Viaducts  & 76.4/63.4 & 75.9/61.5 & 71.4/60.2 \\
 & Vad  & 54.9/0.0 & 57.2/0.0 & 55.8/0.0 \\
\end{tabularx}
\end{table}

\begin{table}[htp]
\centering
\caption{Input size experiment per dataset. Metrics are im.AUROC/pix.F1Max/im.PG2.
\label{tab:input_size_per_dataset}}
\begin{tabularx}{\columnwidth}{l|l|X|X}
Method & Dataset & Size 128 & Size 512 \\
\hline
\patchcoreshort & RIADs & 83.4/18.9/15.9 & 94.4/46.7/56.6 \\
  & BTech & 93.1/41.7/50.9 & 95.9/64.7/75.5 \\
  & VAD & 79.4/0.0/11.8 & 88.7/0.0/21.0 \\
\hline
\reversedistillationshort & RIADs & 86.6/24.2/27.7 & 94.9/51.3/51.1 \\
  & BTech & 93.2/51.3/64.5 & 82.8/60.6/64.0 \\
  & VAD & 81.8/0.0/12.9 & 69.8/0.0/8.8 \\
\hline
\mmr & RIADs & 78.8/24.7/10.4 & 94.9/51.0/50.5 \\
  & BTech & 86.6/41.8/38.9 & 92.8/54.8/50.9 \\
  & VAD & 79.7/0.0/10.2 & 78.9/0.0/10.4 \\
\hline
\csflowshort & RIADs & 77.2/1.3/12.2 & 89.2/4.8/36.5 \\
 & BTech & 91.5/23.0/41.0 & 95.0/29.4/71.1 \\
 & VAD & 76.1/0.0/12.2 & 84.4/0.0/24.3 \\
\hline
\msflowshort & RIADs & 81.3/1.4/17.9 & 91.5/22.8/42.2 \\
 & BTech & 85.6/17.0/58.4 & 90.8/44.0/64.1 \\
 & VAD & 76.7/0.0/10.7 & 75.6/0.0/14.5 \\
\hline
\simplenetshort & RIADs & 77.4/16.6/11.4 & 94.2/40.7/54.2 \\
 & BTech & 91.9/37.7/61.3 & 87.5/38.5/41.3 \\
 & VAD & 61.0/0.0/4.9 & 61.7/0.0/3.9 \\
 \hline
\draemshort & RIADs & 82.5/53.5/23.4 & 84.6/45.5/19.3 \\
 & BTech & 88.9/31.6/49.8 & 90.8/24.4/23.9 \\
 & VAD & 70.1/0.0/7.2 & 59.3/0.0/4.3 \\
\hline
\glassshort & RIADs & 82.9/26.9/18.3 & 92.1/55.1/36.2 \\
 & BTech & 74.5/29.6/32.1 & 95.2/51.0/62.1 \\
 & VAD & 74.8/0.0/11.4 & 80.3/0.0/12.8 \\
\hline
\end{tabularx}
\end{table}

Another interesting outcome in the Noisy Labels experiment is that adding a little label contamination (4\%) strongly improves the separation of bad images (PB2) for almost all models. PatchCore shows the best improvement at 30.2 points, ReverseDistillation improves by 22.8 points. This phenomenon requires further investigation. It also demonstrates the importance of metrics, which qualify the classification results differently for good and bad parts.

Results per dataset are in Tab.~\ref{tab:noisy_labels_per_dataset}, Tab.~\ref{tab:input_size_per_dataset} and Tab.~\ref{tab:data_drift_per_dataset}. Tab.~\ref{tab:input_size_per_dataset} gives a particular insight into how input size is connected to the size of defects presented in the dataset. RIADs show improvement for all models, demonstrating that a bigger input size helps to detect small defects. Meanwhile, VAD and BTech show no improvement or even reduction due to some large defects not processed properly by the feature extractor.

\begin{table}[htp]
\centering
\caption{Data drift experiment per dataset. Metrics are im.AUROC/pix.AUPRO/im.PG2.
\label{tab:data_drift_per_dataset}}
\begin{tabularx}{\columnwidth}{l|l|X|X}
Method & Dataset & No drift        & Drift          \\
\hline
\patchcoreshort & RIADs & 91.4/92.0/39.0 & 65.9/19.5/0.7 \\
 & BTech & 95.5/76.9/67.3 & 87.8/40.4/28.8 \\
 & VAD & 88.0/0.0/16.5 & 69.4/0.0/13.0 \\
\hline
\reversedistillationshort & RIADs & 93.2/95.0/46.3 & 47.9/25.5/0.0 \\
 & BTech & 94.3/79.5/67.7 & 46.2/11.4/4.1 \\
 & VAD & 84.7/0.0/20.1 & 51.7/0.0/2.0 \\
\hline
\mmr & RIADs & 92.4/96.3/42.7 & 67.2/15.4/16.7 \\
 & BTech & 93.7/77.9/62.9 & 67.0/26.7/22.3 \\
 & VAD & 87.6/0.0/27.6 & 61.3/0.0/12.8 \\
\hline
\csflowshort & RIADs & 86.3/47.9/22.7 & 57.9/19.0/0.0 \\
 & BTech & 95.1/57.5/71.5 & 67.9/12.4/8.7 \\
 & VAD & 82.2/0.0/17.1 & 52.4/0.0/2.0 \\
\hline
\msflowshort & RIADs & 89.2/85.6/31.1 & 55.7/7.1/6.7 \\
 & BTech & 90.0/62.6/57.4 & 51.9/0.0/5.9 \\
 & VAD & 84.4/0.0/26.6 & 53.1/0.0/7.2 \\
\hline
\draemshort & RIADs & 85.6/86.8/26.9 & 52.4/16.4/7.2 \\
 & BTech & 89.6/52.6/55.9 & 53.7/9.6/4.8 \\
 & VAD & 57.7/0.0/1.8 & 51.2/0.0/3.1 \\
\hline
\glassshort & RIADs & 88.8/67.1/30.7 & 56.1/15.8/5.8 \\
 & BTech & 90.9/61.1/45.5 & 53.6/7.7/6.4 \\
 & VAD & 79.2/0.0/10.3 & 58.1/0.0/5.1 \\
\hline
\simplenetshort & RIADs & 82.6/77.6/35.4 & 55.1/15.8/0.0 \\
 & BTech & 89.7/68.0/53.5 & 57.3/15.4/6.1 \\
 & VAD & 69.7/0.0/5.3 & 57.1/0.0/4.1 \\
\hline
    \end{tabularx}
    \end{table}

\end{document}